# IMDB Spoiler Dataset

**Rishabh Misra**

UC San Diego

r1misra@eng.ucsd.edu

**Abstract.** User-generated reviews are often our first point of contact when we consider watching a movie or a TV show. However, beyond telling us the qualitative aspects of the media we want to consume, reviews may inevitably contain undesired revelatory information (i.e. 'spoilers') such as the surprising fate of a character in a movie, or the identity of a murderer in a crime-suspense movie, etc. In this paper, we present a high-quality movie-review based spoiler dataset to tackle the problem of spoiler detection and describe various research questions it can answer.

## 1. Motivation

For people who are interested in consuming media but are unaware of the critical plot twists, spoilers may decrease the excitement regarding the pleasurable uncertainty and curiosity of media consumption. A random tweet, review, or recommended news article can greatly spoil people's experience when they have been looking forward to watching something. Therefore, a natural question is how to identify these spoilers in entertainment reviews, so that users can more effectively navigate review platforms. The first step towards solving the problem is access to a high-quality dataset which can highlight spoiler/non-spoiler reviews. Motivated by this utility, we curate the *IMDB Spoiler Dataset* presented in this paper.

## 2. IMDB Spoiler Dataset

We present[1] a large-scale and high-quality spoiler dataset of movie reviews collected from IMDb[2]. When leaving reviews on IMDb, users have the capability to annotate whether their review has any spoilers. Table 1 notes the overall statistics of the dataset. We have a total of 573,913 reviews out of which 150,924 contain spoilers (~26.3%).

---

[1] Dataset is available at https://rishabhmisra.github.io/publications/
[2] https://www.imdb.com/

| Statistic | Value |
| --- | --- |
| # reviews | 573,913 |
| # spoiler reviews | 150,924 |
| # users | 263,407 |
| # movies | 1,572 |
| # users with at least one spoiler review | 79,039 |
| # movies with at least one spoiler review | 1,570 |

Table 1: General statistics of the dataset.

The dataset is divided into 2 files: `IMDB_movie_details` contains metadata about the movies in the dataset and `IMDB_reviews` contains user reviews to the movies available in the dataset.

Each record in `IMDB_movie_details` consists of the following attributes:
- `movie_id`: unique id for the movie.
- `plot_summary`: summary of the plot without any spoilers.
- `duration`: run time of the movie.
- `genre`: list of genres the movie belongs to.
- `rating`: star rating out of 10.
- `release_date`: date the movie was released.
- `plot_synopsis`: revealing details of the plot of the movie.

Each record in `IMDB_reviews` consists of the following attributes:
- `review_date`: date the review was posted.
- `movie_id`: unique id of the movie for which the review is about.
- `user_id`: unique id of the user who left the review.
- `is_spoiler`: whether the review contains spoilers.
- `review_text`: text of the review.
- `rating`: star rating (out of 10) given by the user.
- `review_summary`: review summary provided by the user.

As we can see, apart from the spoiler information, we have included a variety of metadata that can prove useful in tackling various prediction problems apart from spoiler detection. We expand more on this in section 6.

## 3. Data Curation Method

We make use of open-source tools like BeautifulSoup, Selenium, and Chrome Driver to curate the dataset. First, we extracted about 1500 seed movie links from the IMDb page. In all the movies presented, we extract their metadata like id, plot summary, genre information, average rating, and release date using BeautifulSoup API. Next, for each movie, we scraped all the available user reviews along with information about whether the review has spoilers, ratings, and review date. Since not all the reviews are loaded at once, we simulate a "load more" button click using Selenium to load extra reviews and repeat the process until there are no more reviews. Due to this method of collection, we have much more users as compared to movies in the dataset.

## 4. Reading the Data

Once you download the dataset, you can use the following code snippet to read the data for your machine learning methods:

```
import json

def parse_data(file):
    for l in open(file,'r'):
        yield json.loads(l)

data = list(parse_data('./IMDB_reviews.json'))
```

## 5. Exploratory Data Analysis

In order to visualize how reliable the spoiler annotation on IMDb is, we simply visualize the spoiler/non-spoiler annotated reviews that contain the word "spoiler".

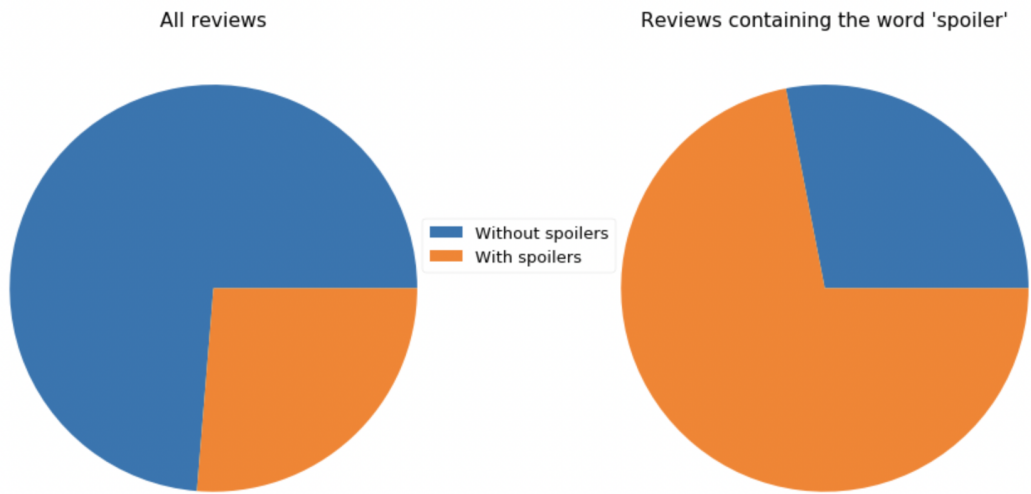

Figure 1: Fraction of spoiler/non-spoiler reviews based on IMDb annotation.

We notice that >25% of times, people may call out a spoiler in their review but may not have annotated it containing spoiler information on IMDb. So there is a 1 in 4 chance that users may accidentally read spoilers while browsing IMDb, which motivates the need to develop sophisticated spoiler detection tools.

In Figure 2, we plot the distribution of word counts for spoiler/non-spoiler reviews and notice they follow a similar pattern, so there is no apparent distinction on the surface.

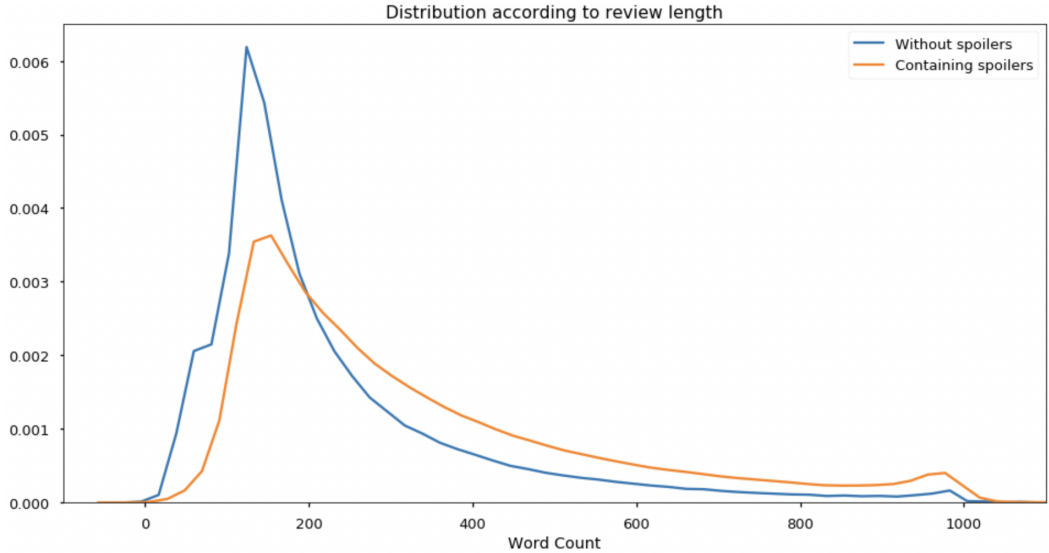

Figure 2: Distribution of spoiler and non-spoiler reviews based on word count.

Next, we take a case study of a specific movie: *Star Wars: Episode V - The Empire Strikes Back (1980)*. Based on the insights produced by Wan et. al., we want to ascertain whether spoiler reviews contain more movie-specific words. *Star Wars: Episode V - The Empire Strikes Back (1980)* has 401 reviews out of which 95 are spoilers (~23.7%). In figure 3, we visualize how the fraction of spoiler reviews increases when looking at some movie-specific words, which validates the findings by Wan et. al.

## 6. Potential Use Cases

Apart from the evident spoiler detection task, the *IMDB Spoiler Dataset* also contains sufficient metadata to study the problem of movie recommendations based on users' reviews and ratings. Furthermore, rich textual information present in the movies combined with genre information can aid in tackling the genre prediction problem, which has utility for understanding text semantics.

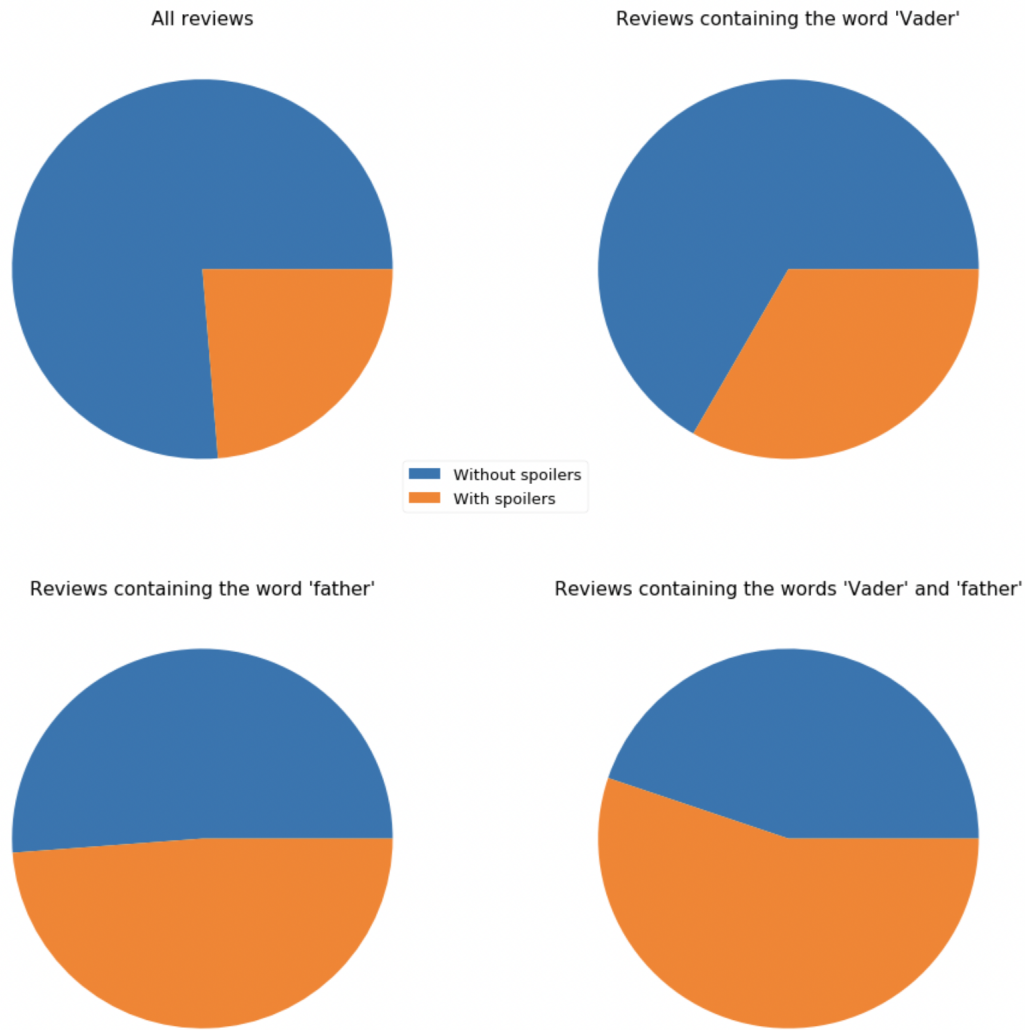

Figure 3: For *Star Wars: Episode V - The Empire Strikes Back (1980)*, from left to right and top to bottom, we plot a fraction of spoiler reviews among the review, among the reviews which mention `Vader`, among the reviews which mention `father` and among the reviews which have both `Vader` and `father`. We notice as movie-specific words increase, the fraction of spoiler reviews also increases.